\documentclass[runningheads]{llncs}

\usepackage[camera-ready]{eccv}

\usepackage{eccvabbrv}

\usepackage{graphicx} 
\usepackage{booktabs} 
\usepackage{multirow}
\usepackage{booktabs}   
\usepackage{makecell}   
\usepackage[table]{xcolor} 

\definecolor{mypink}{RGB}{230, 240, 255} 
\definecolor{mygray}{gray}{0.6}        
\usepackage{url}
\usepackage{booktabs}
\usepackage{amssymb}
\usepackage{bbding}
\usepackage{pifont}
\usepackage{wasysym}
\usepackage{utfsym}
\usepackage{fontawesome}

\usepackage{graphicx}
\usepackage{booktabs}

\usepackage[normalem]{ulem}

\usepackage[accsupp]{axessibility}  
\usepackage{hyperref}

\usepackage{orcidlink}
\usepackage{soul}

\begin{document}

\title{Dynamic-Robust Photometric-Semantic Reconstruction for Open-Vocabulary 3D \\
Scene Understanding} 

\titlerunning{SPAR}

\author{
Boyu Cai$^{1,2}$,
Li Yang$^{2*}$,
Yan Xu$^{3}$,
Wei Liu$^{2}$,
Nian Liu$^{2}$,
Sikui Zhang$^{2}$,
Yan Wang$^{4\ddagger}$,
Chunfeng Yuan$^{2}$,
Weiming Hu$^{1,2}$
}

\authorrunning{B.~Cai et al.}

\institute{School of Information Science and Technology, ShanghaiTech University \and
State Key Laboratory of Multimodal Artificial Intelligence Systems (MAIS), Beijing Key Laboratory of Super Intelligent Security of Multi-Modal Information,
Institute of Automation, Chinese Academy of Sciences (CASIA) \and
Electronic Engineering Department, The Chinese University of Hong Kong \and 
Deepeleph Intelligent Technology
\\
\email{caiby2024@shanghaitech.edu.cn, li.yang@nlpr.ia.ac.cn}}

\maketitle

\begingroup
\renewcommand{\thefootnote}{}
\footnotetext{
$^*$ Corresponding author.
\\
$^\ddagger$ Joint work with Zhejiang Deepeleph Intelligent Technology Co., Ltd.

}
\endgroup

\begin{abstract}
The integration of novel view synthesis (NVS) and open-vocabulary segmentation (OVS) has recently yielded powerful feed-forward 3D foundation models. However, their inherent reliance on static-scene assumptions leads to severe misalignment of spatial features in unconstrained dynamic environments.
To bridge this critical gap, we propose SPAR, a novel joint semantic-geometric encoding architecture that explicitly isolates transient dynamic noise prior to latent space aggregation. Furthermore, we introduce a dynamic-region-aware end-to-end training paradigm that structurally couples motion estimation with multi-view visual and semantic learning. This unified approach enables the network to inherently resolve motion conflicts and distill multi-view consistent, temporally stable scene representations from dynamic inputs. 
Extensive experiments on the challenging D-RE10K benchmark demonstrate that SPAR achieves state-of-the-art performance. 
Our end-to-end approach achieves exceptional novel view synthesis quality, yielding a PSNR of 22.15 dB and 23.33 dB given only 3 and 4 input views respectively. Despite being trained in a self-supervised manner, our model achieves an mIoU of $88.5\%$ for motion mask prediction.
Furthermore, our analysis reveals a strong inter-task synergy between photometric scene reconstruction and semantic understanding, where semantic synthesis learning consistently enhances photometric fidelity in novel view rendering. Code will be available at \url{https://github.com/dmucby/SPAR}.
\end{abstract}    
\section{Introduction}
\label{sec:intro}
Recent advances in novel view synthesis~(NVS)~\cite{3dgs, nerf} and open-vocabulary segmentation~(OVS)~\cite{radford2021clip, lseg} have pushed 3D perception beyond photorealistic rendering toward unified semantic scene understanding. Instead of optimizing a scene-specific language NeRF~\cite{kerr2023lerf} or language Gaussian~\cite{qin2024langsplat}, feed-forward methods such as LSM~\cite{lsm}, SIU3R~\cite{wei2026siu3r} infer geometry, appearance, and semantic representations directly from sparse and unposed images. By combining large reconstruction models~\cite{wang2024dust3r} and open-vocabulary segmentation models~\cite{lseg,cheng2022mask2former}, these methods jointly model geometry and semantics in 3D fields and achieve strong results on benchmarks such as ScanNet~\cite{scannet}.

\begin{figure}[t]
  \centering
  \includegraphics[width=\textwidth]{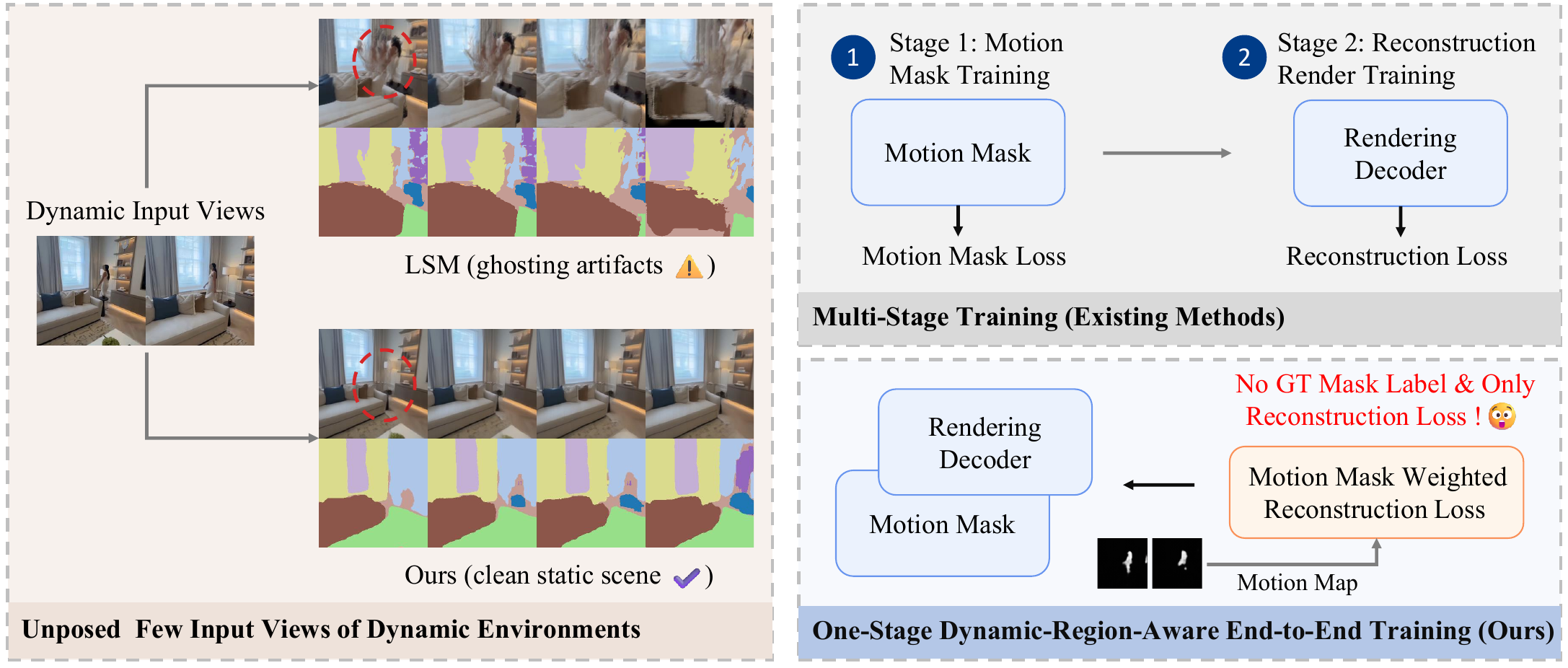}%
  \caption{ Comparison under dynamic, unposed settings. Baselines such as LSM~\cite{lsm} struggle with moving objects, producing ghosting artifacts and inconsistent semantics (left top). By masking dynamic regions prior to latent encoding, our SPAR generates clean, ghost-free photometric and semantic renderings in a single forward pass (left bottom).
}
  \label{fig:teaser}
\end{figure}

However, this paradigm relies on a strict assumption: all input views are expected to depict the same static 3D scene. 
In real-world videos, this assumption is often violated by moving humans, animals, and other transient foreground objects.
Because a moving object may occupy different 3D locations across frames, its corresponding image regions cannot be reconciled within a single static 3D scene representation.
When these regions are aggregated with static content, they introduce conflicting geometry and semantics into the scene representation, leading to ghosting artifacts in RGB rendering and inconsistent semantic predictions (see Figure~\ref{fig:teaser}, left).

To address this limitation, we propose \textbf{SPAR}, a \textbf{S}emantic-\textbf{P}hotometric-\textbf{A}ware \textbf{R}econstruction framework that jointly performs novel view synthesis and semantic understanding from a few unposed observations of dynamic environments. 
SPAR adopts an encoder-decoder architecture that jointly embeds RGB observations and open-vocabulary semantic features into a latent scene representation. 
A shared rendering decoder then queries this representation to synthesize target-view RGB images and semantic feature maps. 
To prevent moving objects from contaminating the latent scene representation, SPAR introduces a Cross-View Dynamic Region Predictor (CV-DRP), which estimates motion masks from cross-view cues and filters dynamic foreground regions before scene-level encoding.


Furthermore, to avoid costly dynamic-objects labeling and maintain scalability, we introduce a dynamic-region-aware end-to-end training paradigm. 
This paradigm couples dynamic region estimation with multi-view photometric and semantic modeling within a unified optimization framework (See Figure~\ref{fig:teaser}, right). 
By using the predicted motion masks to spatially gate reconstruction errors, our approach focuses supervision on multi-view consistent static regions while suppressing gradients from transient foregrounds. As a result, photometric view synthesis and semantic modeling mutually reinforce each another, enabling the network to resolve motion conflicts and recover a clear static 3D scene representation in a single forward pass.


We evaluate SPAR on the challenging D-RE10K benchmark \cite{wildrayzer}, which contains unconstrained indoor sequences with complex motions of transient objects. 
Given only a few input views, SPAR achieves state-of-the-art performance, reaching a PSNR of $22.15$ dB and $23.33$ dB with $3$ and $4$ input views, respectively. 
Notably, our dynamic region predictor achieves an $88.5\%$ mIoU under $3$ input views, surpassing leading self-supervised methods by more than $36$ percentage points.  
Furthermore, ablation studies reveal a strong inter-task synergy: rather than compromising rendering capacity, the semantic branch serves as a structural regularizer that improves the photometric fidelity of the reconstructed scene. 

In summary, our main contributions are as follows:
\begin{itemize}
\item We propose SPAR, the first unified feed-forward framework for joint novel view synthesis and open-vocabulary semantic understanding from a few unposed views of dynamic environments.

\item We design a novel Cross-View Dynamic Region Predictor with a dynamic-region-aware optimization paradigm. 
By using the predicted motion mask to spatially weight both photometric and semantic reconstruction losses, our method enables end-to-end joint training without requiring ground-truth motion mask labels.

\item Extensive experiments on the D-RE10K benchmark demonstrate that SPAR establishes new state-of-the-art performance in both dynamic NVS and motion mask prediction. 
These results validate the mutual benefits of jointly learning photometric reconstruction and semantic scene understanding.
\end{itemize}
\section{Related Work}
\subsection{Generalizable Novel View Synthesis}
Instead of optimizing a scene-specific representation~\cite{nerf,3dgs, xu2026novel,lu2023urban},  generalizable novel view synthesis (NVS) targets fast inference by training across diverse scenes such that novel views (or an intermediate 3D representation) can be predicted in a single forward pass.
Early generalizable methods infer volumetric features from sparse inputs and exploit strong geometric inductive biases, e.g., epipolar constraints and plane-sweep cost volumes, as in PixelNeRF~\cite{yu2021pixelnerf}, MVSNeRF~\cite{chen2021mvsnerf}, and IBRNet~\cite{wang2021ibrnet}.
Subsequent works improve robustness in sparse-view settings and extend feed-forward reconstruction to efficient 3D Gaussian representations, including pixelSplat~\cite{charatan2024pixelsplat} and MVSplat~\cite{chen2024mvsplat}, while pose-free Gaussian inference further reduces reliance on calibrated cameras at test time~\cite{ye2024nopose,li2026noperoomgs}.
More recently, large reconstruction models (LRMs) scale model capacity and data to learn generic 3D priors~\cite{hong2024lrm,zhang2024gslrm,chen2024longlrm, croco}.
In parallel, token-space renderers largely remove hand-crafted 3D inductive biases, exemplified by LVSM~\cite{lvsm}.
RayZer~\cite{rayzer} continues this transformer-based direction and introduces self-supervised, pose-free NVS without 3D supervision, yet it still assumes static imagery.
WildRayZer~\cite{wildrayzer} explicitly lifts this static-scene assumption by learning transient-region masks and gating both tokens and losses, enabling feed-forward NVS under dynamic, in-the-wild inputs.

\subsection{Scene Understanding}
Substantial progress has been made in 2D and 3D scene understanding.
In 2D, vision-language pretraining such as CLIP~\cite{radford2021clip, Yang2026SRADet} enables open-vocabulary recognition, and LSeg~\cite{lseg} aligns dense predictions with language features, 
while object-level detection and grounding have also been extensively studied~\cite{yang2022pdnet,yang2022improving}.
Query-based transformers, e.g., MaskFormer~\cite{cheng2021maskformer} and its universal extensions such as Mask2Former~\cite{cheng2022mask2former}, provide a unified formulation for semantic/instance/panoptic segmentation, and recent advances extend these paradigms to video settings~\cite{zhang2023dvis,zhang2025dvispp,sam2}.
However, 2D understanding is inherently viewpoint-dependent and does not directly enforce cross-view spatial consistency.
For 3D understanding, recent approaches operate on pre-scanned point clouds (often clustered into superpoints) and employ mask/proposal-style formulations~\cite{schult2023mask3d,lu2023qrt,xu2024unified3d,yang2023exploiting}, but they typically require costly 3D acquisition and preprocessing, limiting scalability in practical deployments.

\subsection{Simultaneous Understanding and Novel View Synthesis}
A growing line of work seeks to couple NVS with scene understanding by embedding semantic or language features into 3D representations.
Representative approaches inject foundation-model features into NeRFs~\cite{nerf-ddf,kerr2023lerf} or 3D Gaussians~\cite{zhou2024feature3dgs,qin2024langsplat,zuo2025fmgs}, usually by aligning 3D fields with 2D features through differentiable rasterization.
These pipelines often rely on dense captures and per-scene optimization, which limits efficiency at scale.
LSM~\cite{lsm} mitigates this drawback by performing 2D-to-3D feature alignment within a large feed-forward reconstruction model.
Nevertheless, the feature-alignment paradigm inherits two fundamental limitations: (i) instance-level understanding is constrained by the capabilities of the underlying 2D teacher, and (ii) semantic fidelity can degrade due to aggressive feature compression required by rasterization and memory constraints~\cite{wei2026siu3r}.
SIU3R~\cite{wei2026siu3r} proposes an alignment-free alternative based on explicit pixel-aligned 2D-to-3D lifting and a unified query decoder, enabling native multi-task 3D understanding alongside reconstruction.
Orthogonally, our method highlights that simultaneous modeling must also address dynamics: without explicitly disentangling transient content, multi-view consistency breaks and corrupts both reconstruction and downstream understanding.
Overall, recent evidence suggests that scalable simultaneous understanding and NVS benefit from (a) a joint photometric-semantic render and (b) explicit mechanisms to handle in-the-wild dynamics.
\section{Method}

\subsection{Overview}
\begin{figure}[t]
  \centering
  \includegraphics[width=\textwidth]{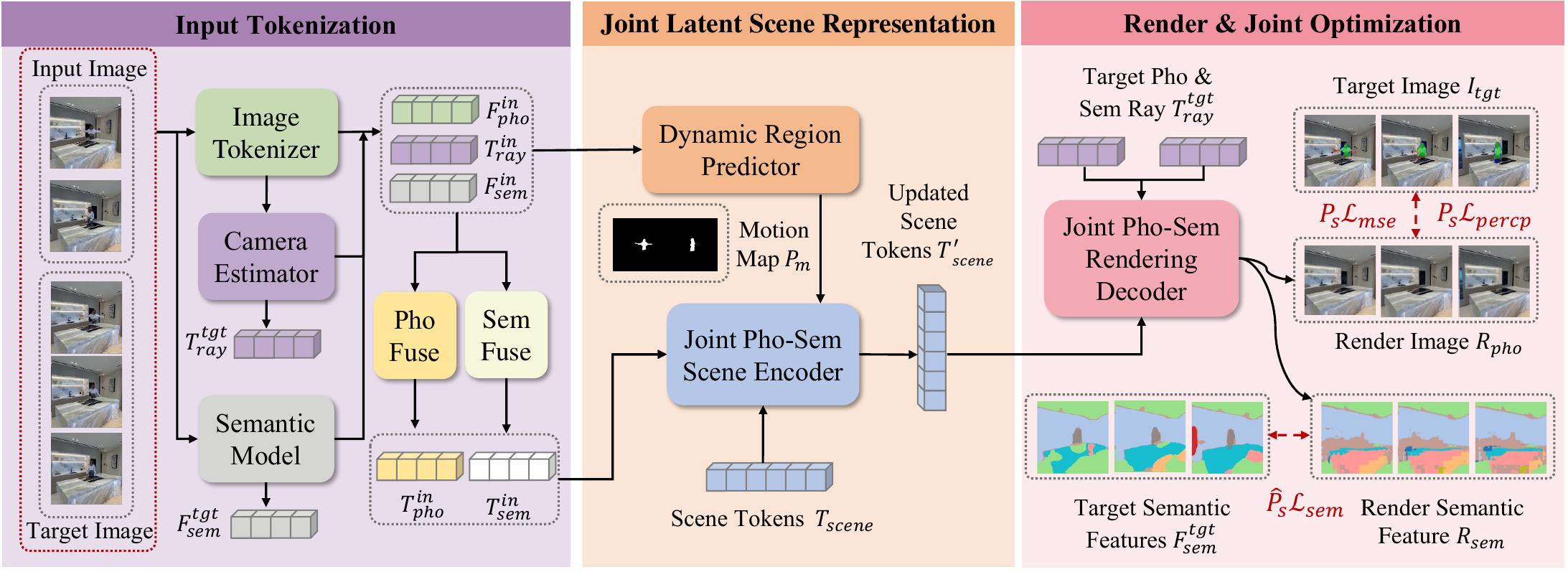}%
  \caption{ 
  Overview of SPAR. Our unified encoder-decoder jointly performs novel view synthesis and open-vocabulary scene understanding from unposed dynamic views. The pipeline first estimates camera poses to build pose-conditioned photometric and semantic tokens. A dynamic region predictor masks moving foregrounds to isolate static geometry. A scene encoder aggregates filtered tokens into a compact latent scene representation, which a dual-head decoder uses to render the target image and semantic features.
}
  \label{fig: framwork}
\end{figure}

We propose \textbf{SPAR}, a unified framework that jointly performs novel view synthesis and semantic understanding from unposed multi-view observations of dynamic environments. 
As illustrated in Figure~\ref{fig: framwork}, SPAR follows a latent scene representation-based encoder-decoder architecture. Given multi-view images, the network first estimates per-image camera poses and constructs pose-conditioned photometric and semantic tokens. To handle dynamic content, a dynamic region predictor identifies the moving foreground regions and masks the associated photometric and semantic tokens to mitigate the impact of motion-induced inconsistencies. The scene encoder then aggregates the masked tokens into a set of scene tokens, forming a compact scene representation that jointly encodes photometric and semantic cues. Given the scene tokens and the target-view pose, the decoder predicts the target view via two parallel heads, producing both the rendered photometric image and semantic feature maps. The semantic features are further aligned with text embeddings to produce open-vocabulary semantic segmentation for the novel views.


Moreover, we design an end-to-end training paradigm that couples the optimization of dynamic region estimation with multi-view visual and semantic learning, enabling SPAR to learn multi-view consistent and temporally stable scene representations from dynamic,  multi-view inputs.



\subsection{Model Architecture}

\subsubsection{Input Tokenization.}
Given multi-view images $I_i \in \mathbb{R}^{H \times W \times 3}$, we follow RayZer \cite{rayzer} to tokenize the images together with their estimated pose Plücker rays into patch-level tokens, which are fused to form the photometric tokens $T_{pho}\in \mathbb{R}^{N_{pho} \times D}$. Meanwhile, we extract the semantic features of the input images using a pre-trained 2D semantic segmentation model (e.g., LSeg~\cite{lseg}). The corresponding pose Plücker rays are also fused with these features to obtain the semantic tokens $T_{sem} \in \mathbb{R}^{N_{sem} \times D}$.



To learn stable scene representations from input views in dynamic environments, we suppress dynamic foreground regions in both the photometric and semantic tokens. Specifically, we employ a dynamic region predictor to estimate the masks of dynamic regions for the input images. These masks are then applied to the photometric and semantic tokens to suppress tokens corresponding to dynamic regions, producing the masked photometric tokens $\tilde{T}_{pho}$ and masked semantic tokens $\tilde{T}_{sem}$.



\subsubsection{Joint Pho-Sem Scene Encoder.} 
The scene encoder aggregates multi-view information from the masked photometric tokens $\tilde{T}_{pho}^{in}$ ($in$ denotes the input views) and semantic tokens $\tilde{T}_{sem}^{in}$, and produces a compact set of scene tokens that capture the global scene representation. 


To this end, we initialize a set of learnable global scene tokens $T_{scene} \in \mathbb{R}^{S \times D}$, and concatenate them with the masked photometric tokens and semantic tokens to form $T_{all} = \text{Concat}(T_{scene}, \tilde{T}_{pho}^{in}, \tilde{T}_{sem}^{in}) \in \mathbb{R}^{(S+N_{pho} + N_{sem}) \times D}$.
$T_{all}$ is then fed into the scene encoder, which consists of $L$ transformer layers with self-attention. This design enables deep cross-modal interaction among the scene, photometric, and semantic tokens, producing a latent representation $T'_{all}$. Finally, we select the first $S$ tokens from $T'_{all}$ to obtain $T'_{scene} \in \mathbb{R}^{S \times D}$, which serves as the compact scene representation encoding both the photometric and semantic information of the environment.
\subsubsection{Joint Pho-Sem Rendering Decoder. } 
Finally, we design a rendering decoder to jointly decode the photometric and semantic representations of the target views from the encoded scene representation.

We first obtain the pose of the target views using the camera pose estimator and compute their corresponding Plücker rays. Two separate linear layers are then applied to the target-view Plücker rays to generate their photometric and semantic query tokens $T_{ray\_pho}^{tgt}\in \mathbb{R}^{M \times D}$ and $T_{ray\_sem}^{tgt} \in \mathbb{R}^{M \times D}$, respectively ($tgt$ denotes the target views). 
We concatenate these target-view query tokens with the scene tokens $T'_{scene}$ produced by the encoder to form $T_{joint} = \text{Concat}(T_{ray\_pho}^{tgt}, T_{ray\_sem}^{tgt}, T'_{scene}) \in \mathbb{R}^{(2M + S) \times D}$. The combined tokens are fed into the rendering decoder, which performs information interaction via self-attention and produces updated tokens $T'_{joint}$.
We then select the first $2M$ tokens from $T'_{joint}$ and split them into two target-view feature sets $\{T_{out\_pho}, T_{out\_sem}\} \in \mathbb{R}^{M \times D}$. A dual-head MLP renderer is applied to separately predict the target-view image and semantic features from $T_{out\_pho}$ and $T_{out\_sem}$. Specifically, the image rendering head focuses on reconstructing high-frequency appearance details, where $\text{MLP}_{pho}$ maps $T_{out\_pho}$ to RGB values, while the semantic rendering head focuses on semantic alignment, where $\text{MLP}_{sem}$ maps $T_{out\_sem}$ to continuous high-dimensional semantic embeddings rather than fixed classification logits:
\begin{equation}
    R_{pho} = \text{MLP}_{pho}(T_{out\_pho}) \in \mathbb{R}^{M \times 3}, 
    \hfill
    R_{sem} = \text{MLP}_{sem}(T_{out\_sem}) \in \mathbb{R}^{M \times C}
\end{equation}
\subsubsection{Cross-View Dynamic Region Predictor.} 
Before feeding the patch-level photometric and semantic tokens into the encoder, we first need to identify dynamic object regions from the input views to reduce their interference with scene modeling. To this end, we design a \textit{Cross-View Dynamic Region Predictor}
that detects moving foreground regions across the multi-view inputs. 

Specifically, for an input view $I_i \in \mathbb{R}^{H \times W \times 3}$, we first concatenate its semantic features $F_{sem}^{(i)}$, photometric tokens $F_{pho}^{(i)}$ and Plücker ray tokens $T_{ray}^{(i)}$, where $\{F_{sem}^{(i)}, F_{pho}^{(i)}, T_{ray}^{(i)}\} \in \mathbb{R}^{N \times D}$, along the channel dimension to form a joint feature representation $\mathcal{F}^{(i)} \in \mathbb{R}^{N \times 3D}$. 
Since a single-view image lacks sufficient references to determine whether a region belongs to a moving object, we leverage multi-view observations to capture cross-view inconsistencies caused by dynamic objects. 
Concretely, the joint features of paired source and reference views $(\mathcal{F}^{(i)}, \mathcal{F}^{(j)})$ are fed into multiple stacked Transformer self-attention layers to model cross-view relationships. 
The resulting features are then passed through an MLP followed by an upsampling layer to produce a dynamic region probability map $P_{m}^{(i)} \in \mathbb{R}^{H \times W}$ with the same spatial resolution as the input image. The dynamic mask $M^{(i)}$ is obtained by thresholding the predicted probability map:
\begin{equation}
M^{(i)} = \mathbb{I}(P_{m}^{(i)} > \tau).
\end{equation}
At inference time, we optionally refine the predicted masks using SAM2~\cite{sam2} to improve mask quality.
The refined mask is then used to more precisely filter patch tokens of the input views before scene encoding.
\subsection{Dynamic-Region-Aware Optimization}
To learn stable photometric and semantic representations from dynamic and unposed multi-view inputs, we introduce a \textit{dynamic-region-aware optimization} framework. The model jointly performs target-view rendering and dynamic region estimation, where the estimated dynamic regions are used to selectively weight supervision signals. This design allows the model to focus learning on geometrically consistent static regions while reducing the influence of cross-view inconsistent foreground motion. 


\subsubsection{Photometric and Semantic Supervision.}
During training, the model first encodes the scene from the given input view images and then predicts the rendered RGB image $R_{pho}$ and semantic features $R_{sem}$ for each target view.
The rendered outputs are supervised by the ground-truth target-view image $I_{tgt}$ and the corresponding semantic feature map $F_{sem}^{tgt}$ extracted by a pre-trained semantic segmentation model. The photometric reconstruction loss is defined as:
\begin{equation}
    \mathcal{L}_{pho} = \| R_{pho} - I_{tgt} \|_2^2 + \lambda\cdot \mathrm{Percep}(R_{pho}, I_{tgt}),
\label{org_pho_loss}
\end{equation}
where $\mathrm{Percep}(\cdot)$ denotes the perceptual loss~\cite{stereo4d, crowdsampling} and $\lambda$ controls its loss weight.
For semantic supervision, we adopt a cosine similarity loss to align the rendered semantic features with the pre-computed semantic embeddings:
\begin{equation}
    \mathcal{L}_{sem} = 1 - \frac{R_{sem} \cdot F_{sem}^{tgt}}
    {\|R_{sem}\|_2\, \|F_{sem}^{tgt}\|_2},
\label{org_sem_loss}
\end{equation}
\subsubsection{Dynamic-Region-Aware Rendering Loss.}
Moving foreground objects introduce cross-view inconsistencies that can degrade scene representation learning.
To mitigate this issue, we employ the dynamic region predictor to estimate a dynamic region probability map $P_m$ for each target view. The corresponding static region probability map is defined as $P_s = 1 - P_m$. 
This static region probability map is used as a spatial weighting mask for the rendering losses, allowing the model to prioritize supervision from static regions. The dynamic-region-aware photometric loss is therefore defined as: 
\begin{equation}
    \mathcal{L}_{pho}^{dyn} = \frac{1}{\sum P_s} \sum P_s \odot \left(\| R_{pho} - I_{tgt} \|_2^2 + \lambda\cdot\mathrm{Percep}(R_{pho}, I_{tgt}) \right).
\end{equation}
For semantic feature supervision, the static region probability map $P_s$ is downsampled to match the resolution of the semantic features, producing $\hat{P}_s$. The semantic loss is then defined as:
\begin{equation}
    \mathcal{L}_{sem}^{dyn} = \frac{1}{\sum \hat{P}_s} \sum \hat{P}_s \odot \left( 1 - \frac{R_{sem} \cdot F_{sem}^{tgt}}{\|R_{sem}\|_2\, \|F_{sem}^{tgt}\|_2} \right).
\end{equation}
To prevent the network from collapsing to a trivial solution that labels the entire image as dynamic, we introduce a regularization term that penalizes excessive dynamic predictions:
\begin{equation}
    \mathcal{L}_{reg} = \frac{1}{HW} \sum_{i=1}^{H} \sum_{j=1}^{W}\text{BCE}(P_{m}^{(i,j)}, 0).
\end{equation}
We also introduce a copy-paste data augmentation strategy. Specifically, objects are cropped from real images and randomly pasted onto static training scenes to simulate moving foreground objects. For such augmented samples, the original (unaugmented) target image $I_{tgt}$ and its semantic features $F_{sem}^{tgt}$ are used to compute the photometric reconstruction loss $\mathcal{L}_{pho}^{cp}$ and semantic loss $\mathcal{L}_{sem}^{cp}$ following Eq.~\eqref{org_pho_loss}
and Eq.~\eqref{org_sem_loss}. The pasted object mask is used as ground-truth supervision for the predicted dynamic region probability map via a binary cross-entropy loss $\mathcal{L}_{mask}$. 
The overall optimization loss of the unified framework is defined as:
\begin{equation}
    \mathcal{L}_{total} = \lambda_{pho}\mathcal{L}_{pho}^{\{dyn, cp\}} + \lambda_{sem}\mathcal{L}_{sem}^{\{dyn, cp\}} + \lambda_{reg}\mathcal{L}_{reg} + \lambda_{mask}\mathcal{L}_{mask}.
\end{equation}

Our training strategy jointly optimizes multi-view rendering and dynamic region prediction. Rendering errors provide adaptive learning signals for identifying cross-view dynamic regions, while the predicted dynamic regions restrict supervision to geometrically consistent static areas. 
This mutually beneficial interaction enables SPAR to learn robust novel view synthesis and semantic understanding under challenging conditions of dynamic scenes and unknown camera poses. 
\section{Experiments}

\subsubsection{Implementation Details.} 
In our joint framework, SPAR is implemented with a Transformer-based architecture. 
The joint scene encoder, camera estimator, CV-DRP, and joint decoder contain 8, 12, 4, and 12 Transformer layers, respectively. 
The scene token capacity is set to $S=768$. 
We train SPAR on $8\times$ A800 GPUs with a cosine learning rate schedule. 
During pre-training, we randomly mask out 10\% of the input photometric tokens in a portion of the training data to simulate the discarding of dynamic regions in end-to-end training stages.
The end-to-end training stage runs for 20K iterations with a learning rate of $2\times10^{-4}$. 
The loss weights are set to $\lambda_{\text{percp}}=0.2$, $\lambda_{\text{sem}}=0.1$, $\lambda_{\text{mask}}=1$, and $\lambda_{\text{reg}}=0.01$. 
All models are trained at $256\times256$ resolution with a $16\times16$ patch size. 
Additional details are provided in the supplementary material.

\subsubsection{Datasets.}
We utilize three datasets for training: the static RealEstate \cite{re10k} dataset, the dynamic D-RE10K \cite{wildrayzer} dataset and the semantically rich ScanNet \cite{scannet} dataset. For a dynamic environment, model evaluation is performed on D-RE10K, which contains 74 Internet-curated indoor sequences featuring moving humans, pets, and vehicles. Each frame is paired with human-verified motion masks, enabling evaluation specifically restricted to transient regions. For semantic segmentation, we follow LSM \cite{lsm} to conduct evaluation on 40 unseen scenes from ScanNet.

\subsubsection{Evaluation Protocol and Metrics.}
Following \cite{rayzer,wildrayzer}, we use masked PSNR \cite{psnr}, SSIM \cite{ssim}, and LPIPS \cite{lpips} to evaluate the quality of novel view synthesis. To assess the quality of dynamic region predictions, we employ mIoU and Recall for motion mask evaluation. 
The inputs for all evaluations are unposed multi-view images, 
specifically 2, 3, or 4 input views, with 6, 5, or 4 target views, respectively.
Moreover, our model can accept flexible dynamic multi-view inputs at test time, including 2, 3, and 4 input views. 
For semantic segmentation, we report mIoU and Accuracy following LSM~\cite{lsm}, using two input views and one synthesized target view.


\subsection{Main Results}

\begin{table*}[htbp]
\centering
\caption{Performance comparison on D-RE10K under view
settings ($n=2, 3, 4$).}
\label{tab:results}
\small
\resizebox{\columnwidth}{!}{
\begin{tabular}{lccccccccc}
\toprule
\multirow{3}{*}{\textbf{Method}} & \multicolumn{9}{c}{\textbf{D-RE10K}} \\ \cmidrule{2-10} 
 & \multicolumn{3}{c}{\textbf{Views = 2}} & \multicolumn{3}{c}{\textbf{{Views = 3}}} & \multicolumn{3}{c}{\textbf{Views = 4}} \\ \cmidrule(lr){2-4} \cmidrule(lr){5-7} \cmidrule(lr){8-10}
 & PSNR$ \uparrow$ & SSIM$ \uparrow$ & LPIPS$ \downarrow$ & PSNR$ \uparrow$ & SSIM$ \uparrow$ & LPIPS$ \downarrow$ & PSNR$ \uparrow$ & SSIM$ \uparrow$ & LPIPS$ \downarrow$ \\ \midrule
\multicolumn{6}{r}{\textit{Optimization-Based Methods}} \\ \midrule
NeRF On-the-go~\cite{nerfon-the-go} & 15.90 & 0.518 & 0.582 & 18.45 & 0.624 & 0.446 & 19.52 & 0.620 & 0.443 \\
3DGS~\cite{3dgs} & 13.49 & 0.442 & 0.605 & 14.92 & 0.514 & 0.531 & 16.28 & 0.552 & 0.490 \\
T-3DGS~\cite{t-3dgs} & 15.90 & 0.518 & 0.582 & 18.45 & 0.624 & 0.446 & 18.70 & 0.613 & 0.454 \\
Spotless-Splats~\cite{spotlesssplats} & 16.45 & 0.548 & 0.468 & 17.77 & 0.600 & 0.394 & 18.05 & 0.608 & 0.390 \\
WildGaussians~\cite{wildgaussians} & 16.12 & 0.512 & 0.624 & 17.76 & 0.577 & 0.588 & 18.11 & 0.597 & 0.574 \\ \midrule
\multicolumn{6}{r}{\textit{Feed-forward Methods}} \\ \midrule
RayZer + Co-Seg~\cite{co-seg} & 16.76 & 0.547 & 0.516 & 17.53 & 0.565 & 0.580 & 18.67 & 0.636 & 0.481 \\
RayZer + MegaSAM~\cite{megasam} & -- & -- & -- & 20.13 & 0.688 & 0.336 & 20.66 & 0.702 & 0.310 \\
RayZer + SAV~\cite{segment_any_motion} & 19.01 & 0.628 & 0.397 & 20.35 & 0.696 & 0.332 & 20.73 & 0.711 & 0.308 \\
SRT~\cite{srt} & 14.62 & 0.308 & 0.647 & 14.76 & 0.310 & 0.639 & 14.81 & 0.329 & 0.632 \\
LSM~\cite{lsm} & 10.92 & 0.271 & 0.636 & 11.48 & 0.307 & 0.638 & 11.42 & 0.309 & 0.644 \\
SIU3R~\cite{wei2026siu3r} & 13.01 & 0.372 & 0.577 & 13.22 & 0.362 & 0.564 & 13.25 & 0.363 & 0.553 \\
WildRayZer~\cite{wildrayzer} & \textbf{21.78} & \textbf{0.734} & \textbf{0.308} & \underline{21.98} & \textbf{0.754} & \underline{0.314} & \underline{22.38} & \textbf{0.773} & \underline{0.290} \\ 
WildRayZer\textsuperscript{\dag}~\cite{wildrayzer} & 19.47 & 0.657 & 0.318 & - & - & - & - & - & - \\ 
Ours & \underline{19.97} & \underline{0.627} & \underline{0.339} & \textbf{22.15} & \underline{0.702} & \textbf{0.283} & \textbf{23.33} & \underline{0.739} & \textbf{0.263} \\ 
\bottomrule
\end{tabular}
}
\end{table*}

\begin{figure*}[t]
  \centering
  \includegraphics[width=\textwidth]{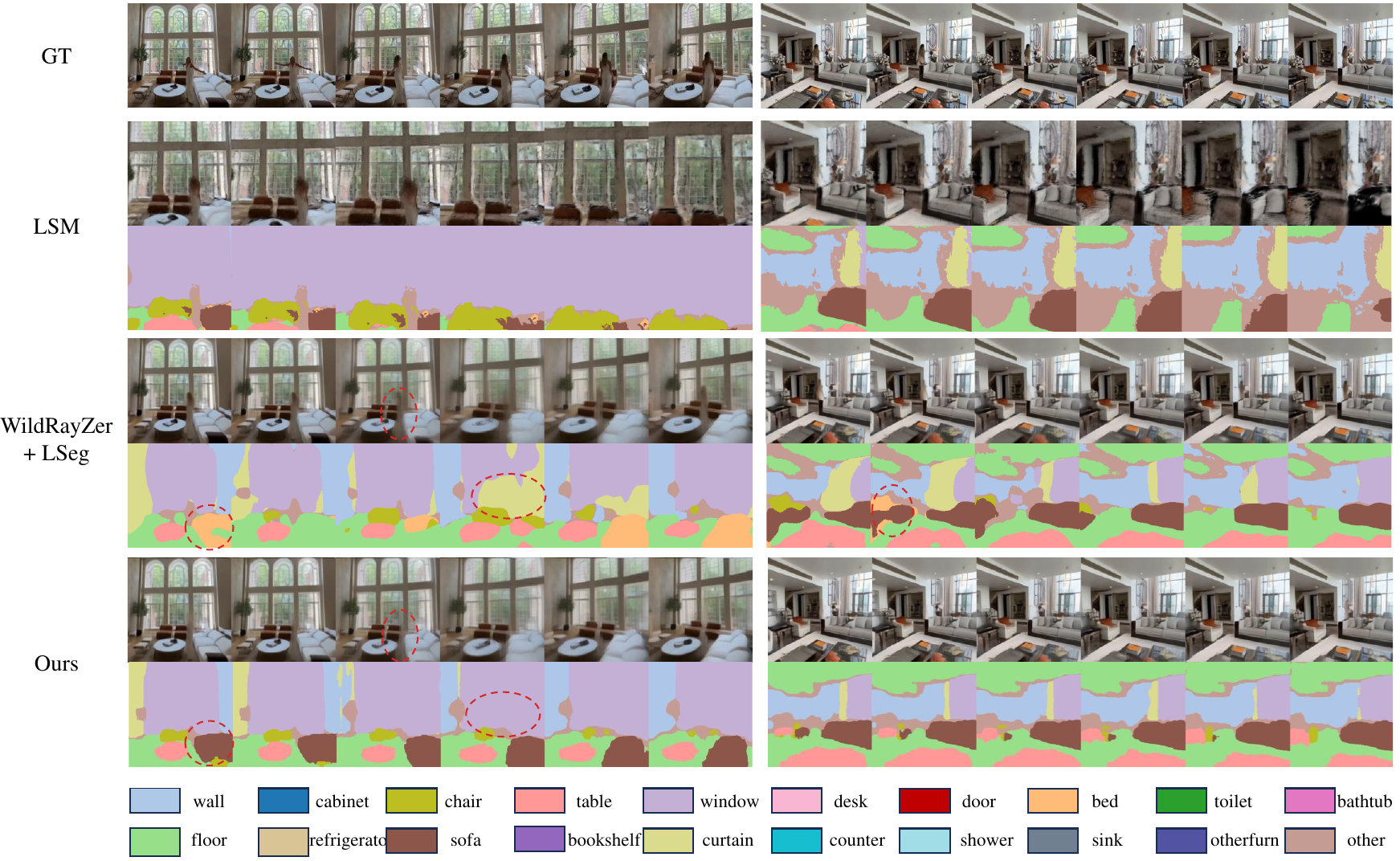}%
  \caption{ 
Qualitative comparison of novel view synthesis and semantic segmentation on dynamic scenes.}
  \label{fig: qualitative}
\end{figure*}

\subsubsection{Novel View Synthesis in Dynamic Environments.}
Table 1 reports the quantitative comparison on D-RE10K under different input-view settings.
With 4 input views, SPAR achieves the best PSNR and LPIPS among all compared methods, reaching 23.33 dB and 0.263, respectively.
The same advantage is observed for 3 input views, where SPAR improves PSNR from 21.98 to 22.15 dB and reduces LPIPS from 0.314 to 0.283 compared with WildRayZer.
In the more challenging 2-view setting, SPAR is below the WildRayZer result reported in the original paper, but achieves a higher PSNR than our reproduction of WildRayZer using the authors' released code and checkpoints.
These results show that SPAR delivers strong feed-forward reconstruction quality for dynamic novel view synthesis, especially when three or more input views are available.

\begingroup
\renewcommand{\thefootnote}{\dag}
\footnotetext{The result is obtained by running the released code and checkpoint. }
\endgroup

\subsubsection{Qualitative Results for Photometric-Semantic Novel view Synthesis.}
We evaluate our single-stage joint semantic and photometric rendering framework against the static-only LSM~\cite{lsm} and the two-stage dynamic rendering approach, WildRayZer~\cite{wildrayzer} + LSeg~\cite{lseg}. As illustrated in Figure~\ref{fig: qualitative}, LSM lacks explicit support for dynamic scenes. It fails in moving regions and produces severely blurred renderings with corrupted segmentation maps. 
The two-stage WildRayZer + LSeg pipeline performs better in dynamic rendering, but its semantic prediction is limited by the quality of the intermediate rendered image. As a result, rendering artifacts can directly degrade segmentation accuracy. For example, it misclassifies the window as a curtain in the left sequence and the sofa as a bed in the right sequence.
In contrast, our method jointly fuses multi-view photometric and semantic features in a unified single-stage rendering process. This design helps reduce the above errors and enables both high-fidelity novel view synthesis and structurally coherent semantic segmentation.

\begin{table}[htbp]
\centering
\small
\caption{\textbf{Dynamic Region quality}. Comparison of supervised and
self-supervised motion segmentation methods under view
settings ($n=2, 3, 8$). 
w/ Refine denotes the use of SAM2 to refine the original motion masks.
}
\label{tab:motion_mask}
\setlength{\tabcolsep}{3.5pt} 
\label{tab:segmentation_results}
\begin{tabular}{l | cc | cc | cc}
\toprule
\multirow{2}{*}{\textbf{Method}} & \multicolumn{2}{c|}{\textbf{$n=2$}} & \multicolumn{2}{c|}{\textbf{$n=3$}} & \multicolumn{2}{c}{\textbf{$n=8$}} \\
\cmidrule(lr){2-3} \cmidrule(lr){4-5} \cmidrule(lr){6-7}
 & mIoU $\uparrow$ & Recall $\uparrow$ & mIoU $\uparrow$ & Recall $\uparrow$ & mIoU $\uparrow$ & Recall $\uparrow$ \\
\midrule
\multicolumn{7}{c}{\textit{Supervised Methods}} \\
\midrule
MegaSAM~\cite{megasam} & - & - & 12.7 & 37.1 & 35.4 & 60.7 \\
Segment Any Motion~\cite{segment_any_motion} & 31.9 & 47.2 & 41.2 & 57.1 & 50.9 & 70.8 \\
Ours w/ Refine & \textbf{87.8} & \underline{84.7} & \textbf{88.5} & \underline{86.2} & \textbf{88.3} & 85.6 \\
\midrule
\multicolumn{7}{c}{\textit{Self-supervised Methods}} \\
\midrule
Co-segmentation~\cite{co-seg} & 9.6 & 45.0 & 13.7 & 53.2 & 16.3 & 45.5 \\
WildRayZer~\cite{wildrayzer} & 53.9 & 85.1 & 52.1 & 84.3 & 54.2 & \textbf{87.7} \\
Ours w/o Refine &  \underline{75.9} & \textbf{85.7} & \underline{75.7} & \textbf{86.4} & \underline{76.6} & \underline{86.6} \\
\bottomrule
\end{tabular}
\end{table}

\begin{figure}[t]
  \centering
  \includegraphics[width=\textwidth]{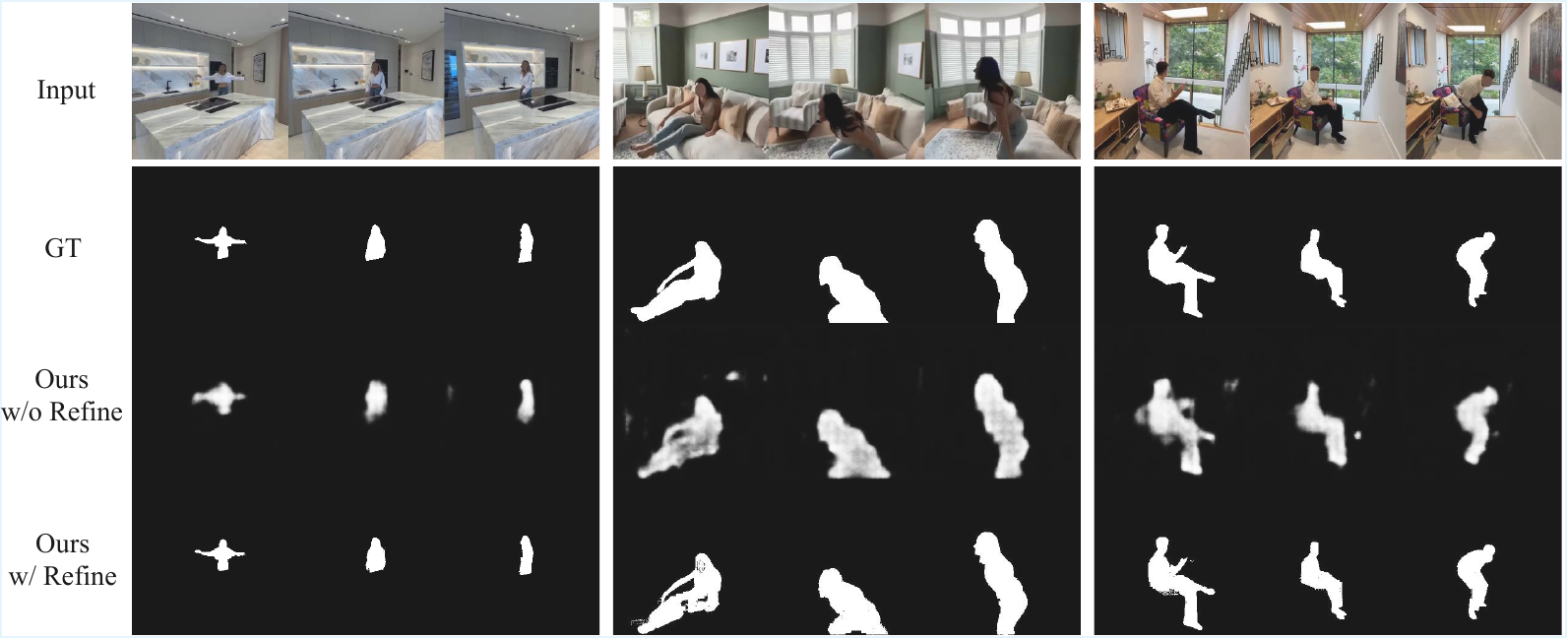}%
  \caption{ 
Qualitative results of our Dynamic Region Predictor (DRP) on diverse multi-view sequences.}
  \label{fig: motion_mask}
  \vspace{-0.8em}
\end{figure}

\begin{figure*}[htbp]
  \centering
  \includegraphics[width=\textwidth]{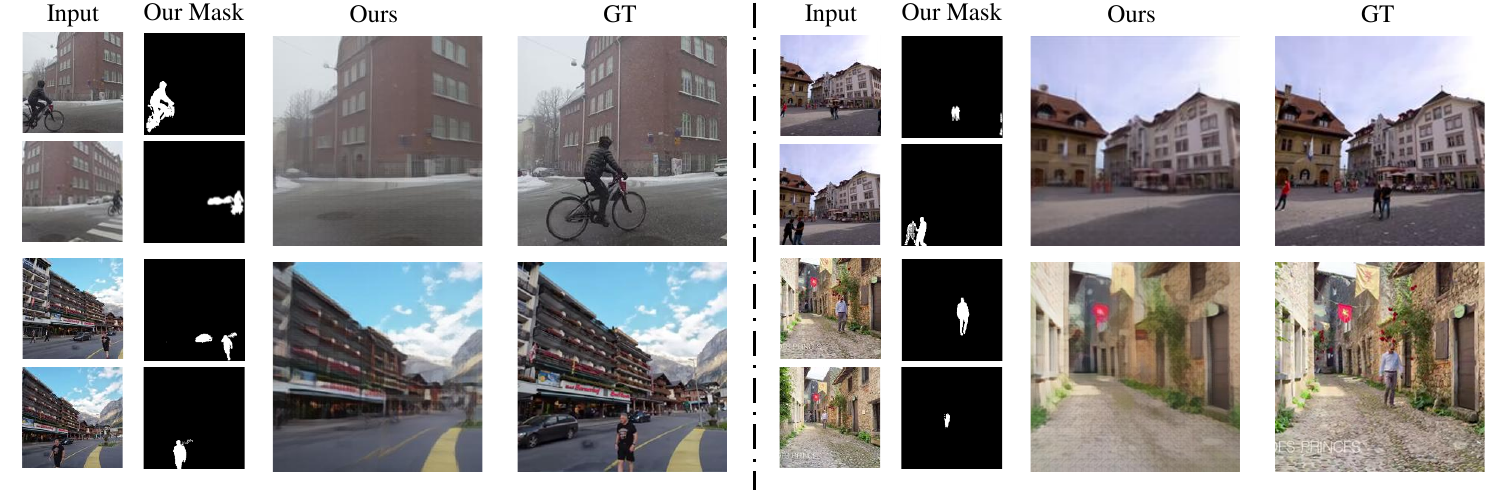}%
  \caption{ 
Qualitative Results of novel view synthesis and dynamic region prediction on unbounded scenes.}
  \label{fig: qualitative_new_data}
  \vspace{-0.8em}
\end{figure*}

\subsubsection{Motion Mask Quality Comparison.} 
As shown in Table~\ref{tab:motion_mask}, SPAR achieves state-of-the-art performance in dynamic region estimation.
We report two variants of our method. \textit{w/o Refine} directly uses the raw dynamic masks predicted by our Dynamic Region Predictor. \textit{w/ Refine} further applies SAM2 only at test time to refine these predicted masks. The two variants use the same trained model, and SAM2 is not involved in training.
Our method consistently outperforms both supervised and self-supervised baselines. Under the $n=3$ setting, Ours w/ Refine reaches 88.5\% mIoU, outperforming the leading self-supervised method WildRayZer~\cite{wildrayzer} by a large margin. Moreover, compared with fully supervised motion-segmentation methods~\cite{segment_any_motion}, our self-supervised framework still achieves better mIoU under the $n=8$ setting.
Notably, our Dynamic Region Predictor is trained without any ground-truth motion mask labels. Under reconstruction-driven optimization, regions that are hard to explain by static multi-view consistency tend to be identified as dynamic, allowing the model to reduce reconstruction errors caused by motion interference.


\subsubsection{Motion Mask Qualitative Results.}
Figure~\ref{fig: motion_mask} visualizes our Dynamic Region Predictor across diverse indoor multi-view sequences. The top row displays the input frames, followed by the ground-truth (GT) masks. 
The third row shows our predictions, denoted as Ours \textit{w/o Refine}. These predictions generate accurate probabilistic heatmaps that reliably localize dynamic human subjects. They remain effective even during complex non-rigid motions, such as sitting and turning. 
The bottom row presents the refined results, denoted as Ours \textit{w/ Refine}. After refinement, these soft spatial priors are converted into precise binary masks that tightly align with the ground truth. This demonstrates the capability of our module as an efficient and robust motion predictor.

\subsubsection{Qualitative Results for Novel View Synthesis and Dynamic Region Prediction on Unbounded Scenes.} 
We directly evaluate our model on videos from SpatialVID~\cite{spatialvid} without any training or fine-tuning on this dataset. 
SpatialVID contains large-scale, open-world, primarily outdoor dynamic scenes with pedestrians, cyclists, vehicles, occlusions, viewpoint changes, and diverse scene layouts. 
As shown in Figure~\ref{fig: qualitative_new_data}, these scenes pose substantial challenges for novel view synthesis due to low visibility, moving foreground objects, occlusions, and unbounded outdoor layouts.
Despite these challenges, our method predicts coherent dynamic-region masks across input views and synthesizes plausible novel views. 
The predicted masks localize moving objects such as cyclists and pedestrians, while the rendered images preserve the overall scene structure. 
These results suggest that SPAR can generalize to unseen outdoor videos in a zero-shot manner.

\subsubsection{Performance Comparisons on the ScanNet Dataset.}
As shown in Table~\ref{tab:scanaet_results}, our method achieves state-of-the-art performance for 3D-aware joint semantic segmentation and novel view synthesis on the semantic-rich ScanNet dataset.
Our semantic branch predicts continuous LSeg/CLIP-aligned embeddings instead of fixed 20-class logits. Therefore, it can inherit the open-set recognition capability of LSeg/CLIP.
For semantic segmentation, LSeg retains a marginal advantage in mIoU by directly segmenting the ground-truth target images, while our method achieves a highly competitive mIoU of 0.5271 and the highest accuracy of 0.8061 under joint rendering. 
For novel view synthesis, our formulation shows clear photometric advantages over existing baselines.
We attribute this improvement to the way we integrate semantic information. Instead of compressing semantic features into individual 3D Gaussians as in LSM~\cite{lsm}, our method embeds uncompressed semantics directly into global scene features, which helps reduce information loss.

\begin{table*}[htbp]
\centering
\small
\caption{Quantitative comparison on the semantically rich ScanNet \cite{scannet} dataset.}
\label{tab:scanaet_results}
\setlength{\tabcolsep}{4.5pt}
\begin{tabular}{l | c c | c c c}
\toprule
\multirow{2}{*}{\textbf{Method}} & \multicolumn{5}{c}{\textbf{Target View}} \\
\cmidrule(lr){2-6}
& mIoU $\uparrow$ & Acc. $\uparrow$ & PSNR $\uparrow$ & SSIM $\uparrow$ & LPIPS $\downarrow$ \\
\midrule
LSeg~\cite{lseg}         & \textbf{0.5281} & 0.7612 & -               & -               & -               \\
NeRF-DFF~\cite{nerf-ddf}     & 0.4037          & 0.6755             & 19.86           & 0.6650          & 0.3629          \\
Feature-3DGS~\cite{zhou2024feature3dgs} & 0.4223          & 0.7174             & 24.49 & \underline{0.8132} & 0.2293 \\
pixelSplat~\cite{charatan2024pixelsplat}   & -               & -                  & 24.89  & \textbf{0.8392} & \underline{0.1641} \\
SRT~\cite{srt}        & -               & - & 17.01          & 0.4130          & 0.6350   \\ 
LSM~\cite{lsm}        & 0.5078 & \underline{0.7686} & 24.39           & 0.8072          & 0.2506   \\ 
SIU3R~\cite{wei2026siu3r}        & -               & - & \underline{25.10}           & 0.8121          & \textbf{0.1043}   \\  
Ours      & \underline{0.5271} & \textbf{0.8061} & \textbf{26.43}           & 0.8052          & 0.2407      \\
\bottomrule
\end{tabular}
\vspace{-0.8em}
\end{table*}

\subsection{Ablation Studies}

\subsubsection{Effectiveness of Dynamic-Region-Aware Optimization.}
As detailed in Table~\ref{tab:optimization_ablation}, the proposed dynamic-region-aware optimization improves robustness under severe dynamic interference. Random masking only provides limited regularization, while our targeted formulation brings a much larger gain. This shows that dynamic foreground estimation provides useful spatial guidance for separating dynamic interference and learning stable static scene representations.


\begin{table}[htbp]
\centering
\small
\caption{\textbf{Ablation of Optimization Strategies}. Quantitative comparison of novel view synthesis performance under severe dynamic interference.}
\label{tab:optimization_ablation}
\setlength{\tabcolsep}{6pt}
\resizebox{\columnwidth}{!}{
\begin{tabular}{l|ccc}
\toprule
\textbf{Optimization Strategy} & PSNR$ \uparrow$ & SSIM$ \uparrow$ & LPIPS$ \downarrow$ \\
\midrule
Unconstrained Baseline                 & 15.66             & 0.489             & 0.532                \\
Baseline + Random Masking              & 17.10 (+1.44)     & 0.536 (+0.047)    & 0.485 (-0.047)       \\
\textbf{+ Dynamic-Region-Aware Opt. (Ours)} & \textbf{19.97} (+4.31) & \textbf{0.627} (+0.138) & \textbf{0.339} (-0.193) \\
\bottomrule
\end{tabular}
}
\vspace{-0.8em}
\end{table}


\subsubsection{Effectiveness of Semantic Branch.} 
As shown in Table~\ref{tab:semantic_branch_ablation}, adding the parallel semantic understanding branch consistently improves NVS performance on D-RE10K.
Compared with the geometry-only baseline, the full unified framework achieves better results across all metrics. This indicates that joint semantic-geometric optimization benefits image rendering and does not compromise reconstruction fidelity.

\begin{table*}[htbp]
\centering
\small
\caption{\textbf{Effectiveness of Semantic Branch}. Quantitative comparison demonstrating the performance gains after incorporating the semantic branch.}
\label{tab:semantic_branch_ablation}
\setlength{\tabcolsep}{6pt}
\begin{tabular}{l | c c c}
\toprule
\textbf{Method} & PSNR$ \uparrow$ & SSIM$ \uparrow$ & LPIPS$ \downarrow$ \\
\midrule
w/o semantic branch & 19.77 & 0.615 & 0.356 \\
w/ semantic branch (ours) & \textbf{19.97} (+0.20) & \textbf{0.627} (+0.012) & \textbf{0.339} (-0.017) \\
\bottomrule
\end{tabular}
\vspace{-0.8em}
\end{table*}

\subsubsection{Ablation of Auxiliary Components.} 
We provide additional ablations by separately removing pre-training, copy-paste augmentation~(CPA), and SAM2 refinement. 
Removing any of these components leads to performance degradation, showing that they all contribute to the final performance. 
Among them, pre-training has the largest impact, indicating that a strong feed-forward reconstruction prior is important for dynamic view NVS. 
In contrast, removing CPA or test-time SAM2 refinement only causes a minor drop. 
This suggests that the improvement is not mainly driven by external mask augmentation or SAM2 post-processing. Instead, our dynamic-region-aware loss in Eq.~(5)--(7) plays the primary role in guiding the model to handle dynamic interference during NVS.

\begin{table}[htbp]
\centering
\small
\caption{\textbf{Ablation of Auxiliary Components}. We ablate pre-training, copy-paste augmentation~(CPA), and SAM2 refinement. }
\label{tab:other_ablation}
\setlength{\tabcolsep}{6pt}
\begin{tabular}{l | c c c}
\toprule
\textbf{Method} & PSNR$ \uparrow$ & SSIM$ \uparrow$ & LPIPS$ \downarrow$ \\
\midrule
Ours w/o pre-training & 18.66 & 0.578 & 0.381  \\
Ours w/o CPA & 19.82 & 0.625 & 0.348 \\
Ours w/o SAM2 refine & 19.90 & 0.624 & 0.341 \\
Ours & \textbf{19.97} & \textbf{0.627} & \textbf{0.339}  \\
\bottomrule
\end{tabular}
\vspace{-0.8em}
\end{table}


\section{Conclusion}
In this paper, we present SPAR, a unified feed-forward framework that alleviates the strict static-scene assumptions limiting many 3D foundation models. SPAR jointly performs novel view synthesis and open-vocabulary semantic understanding from a few unposed dynamic views, addressing spatial feature misalignment caused by transient moving objects. Our method introduces a joint semantic-geometric encoding architecture with a Cross-View Dynamic Region Predictor that suppresses dynamic regions before latent scene aggregation. Combined with a self-supervised dynamic-region-aware training scheme, SPAR encourages optimization to focus on multi-view consistent static regions. This training scheme  relies only on reconstruction loss and does not require ground-truth motion mask labels.

Extensive evaluations on the challenging D-RE10K benchmark show that SPAR achieves state-of-the-art performance in both dynamic NVS and motion mask prediction under extreme few-view settings. Beyond these gains, we observe strong inter-task synergy: semantic learning acts as a structural regularizer that improves photometric reconstruction quality. By enabling robust and temporally stable 3D scene representations without explicit dynamic object annotations, SPAR provides an efficient and scalable foundation for embodied perception in complex real-world environments.



\section*{Acknowledgement}
This work was supported by the National Natural Science Foundation of China (Grant No. 62403462, 62192782, 62532015, U22B2056, 62422317, 62202469), the Key Research and Development Program of Xinjiang Uyghur Autonomous Region (Grant No. 2023B03024), the Beijing Natural Science Foundation (Grant No. L223003, L243015), and the Beijing Major Science and Technology Project (Contract No. Z251100008425008). We thank the authors of LVSM~\cite{lvsm}, RayZer~\cite{rayzer}, WildRayZer~\cite{wildrayzer} for clarifying details of their models. 

%
%
\bibliographystyle{splncs04}
\bibliography{main}


\end{document}